\documentclass[runningheads]{llncs}
\usepackage{graphicx}
\usepackage{svg}
\usepackage{hyperref}
\usepackage{tikz}
\usepackage{comment}
\usepackage{subcaption}
\usepackage{amsmath,amssymb} 
\usepackage{color}

\usepackage[accsupp]{axessibility}  

\usepackage[capitalize]{cleveref}
\usepackage{multirow}

\begin{document}
\pagestyle{headings}
\mainmatter
\def\ECCVSubNumber{4269}  

\newcommand{\todo}[1]{\textcolor{red}{TODO: #1}}

\title{Aware of the History: Trajectory Forecasting with the Local Behavior Data}

\titlerunning{Aware of the History: Trajectory Forecasting with the Local Behavior Data}
%
\author{Yiqi Zhong\inst{1}\orcidID{0000-0002-0928-8018} \and
Zhenyang Ni\inst{2}\orcidID{0000-0001-7134-620X} \and 
Siheng Chen\inst{2,\footnotemark[1]}\orcidID{0000-0001-6199-529X}\and 
Ulrich Neumann\inst{1,\thanks{Corresponding authours}}\orcidID{0000-0001-8977-7112}}
\authorrunning{Y. Zhong et al.}
%
\institute{Department of Computer Science, University of Southern California, Los Angeles, CA 90089, USA \email{\{yiqizhon,uneumann\}@usc.edu}
\and
Cooperative Medianet Innovation Center (CMIC), Shanghai Jiao Tong University, Shanghai, China\\
\email{\{0107nzy,sihengc\}@sjtu.edu.cn}}

\maketitle

\begin{abstract}
The historical trajectories previously passing through a location may help infer the future trajectory of an agent currently at this location. Despite great improvements in trajectory forecasting with the guidance of high-definition maps, only a few works have explored such local historical information. In this work, we re-introduce this information as a new type of input data for trajectory forecasting systems: the \textit{local behavior data}, which we conceptualize as a collection of location-specific historical trajectories. \textit{Local behavior data} helps the systems emphasize the prediction locality and better understand the impact of static map objects on moving agents. We propose a novel local-behavior-aware (LBA) prediction framework that improves forecasting accuracy by fusing information from observed trajectories, HD maps, and local behavior data. Also, where such historical data is insufficient or unavailable, we employ a local-behavior-free (LBF) prediction framework, which adopts a knowledge-distillation-based architecture to infer the impact of missing data. Extensive experiments demonstrate that upgrading existing methods with these two frameworks significantly improves their performances. Especially, the LBA framework boosts the SOTA methods' performance on the nuScenes dataset by at least 14\% for the $K=1$ metrics. \footnote{Code is at \href{ https://github.com/Kay1794/LocalBehavior-based-trajectory-prediction}{https://github.com/Kay1794/LocalBehavior-based-trajectory-prediction}}


\end{abstract}



\section{Introduction}\label{sec:intro}
\begin{figure}[h]
    \includegraphics[width=\textwidth]{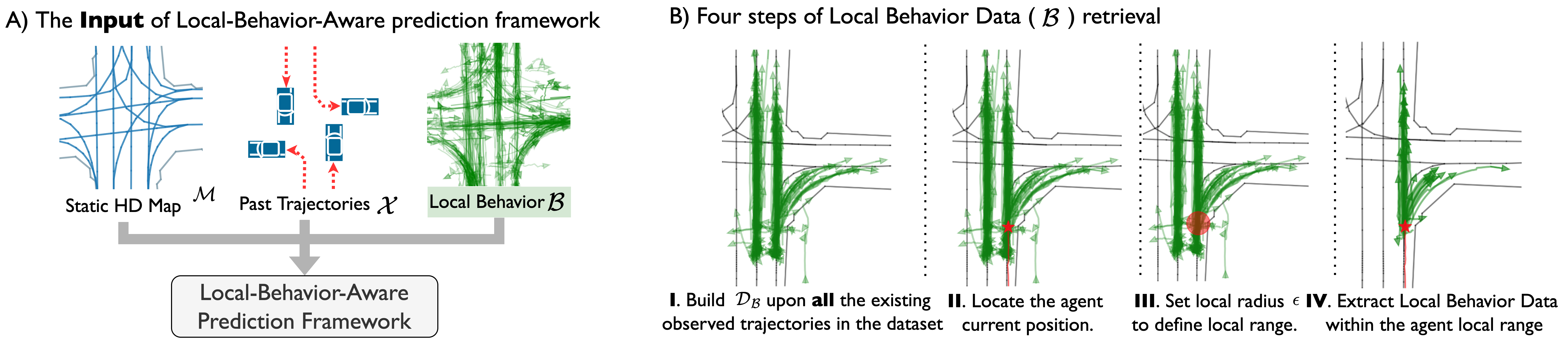}
    
    \caption{A) Compared to previous works that mostly rely on HD maps and agents' past trajectories, we additionally input the local behavior data to the prediction framework. B) For agents to be predicted in the scene, we follow these steps to retrieve their local behavior data for the framework input composition.}
    \label{fig:teaser}
\end{figure}

Trajectory forecasting aims to predict an agent's future trajectory based on its past trajectory and surrounding scene information. This task is essential in a variety of applications, including self-driving cars~\cite{luo2018porca}, surveillance systems~\cite{de2013machine}, robotics~\cite{bennewitz2002learning}, and human behavior analysis~\cite{sun2019relational}. Prior prediction systems primarily use deep-learning-based models (e.g., LSTM, temporal convolutions) to exploit limited information such as past trajectories~\cite{alahi2016social,kim2017probabilistic,nikhil2018convolutional}. Recent efforts also reveal that forecasting ability will improve as more scene information is introduced into the input. One type of scene information, for example, is the past trajectories of a target agent's neighboring agents. To date, many graph-neural-network-based methods have explored the potential of agents' motion features and interactive motion relations to improve predictions~\cite{hu2020collaborative,sun2019stochastic,yu2020spatio}. Recently, high-definition (HD) maps are incorporated as an additional type of scene information~\cite{deo2020trajectory,gao2020vectornet,gilles2021home,liang2020learning,ye2021tpcn} to provide geometric priors.

Besides the widely used HD maps and agents' past trajectories, we propose a novel Local-Behavior-Aware (LBA) prediction framework that takes a new type of scene information as the input, which we term as \emph{local behavior data}. The local behavior data is defined as a collection of historical trajectories at an agent's current location. Fig. \ref{fig:teaser} (A) shows the three components of the LBA framework input. Taking local behavior data as the input brings two benefits to the task. First, the data provides location-specific motion patterns, which helps the model effectively narrow down the search space for future trajectories. Most of the existing prediction models solely rely on the features learned from the static HD map to infer such information~\cite{gu2021densetnt,deo2021multimodal,gilles2021home}. In comparison, taking local behavior data as the input immediately equips the model with this information, making the model more tractable and robust. Second, local behavior data provides complementary information to augment static maps into dynamic behavioral maps. The map prior in the current literature is limited to static geometric information. The rich dynamic information brought by this new input would help the model better understand the impact of map objects on moving agents.


Many car companies and navigation apps are collecting such local behavior data. Yet, sometimes this data is insufficient or is yet to be gathered (e.g. when a self-driving car explores new areas). Therefore, we further propose a Local-Behavior-Free (LBF) prediction framework that only takes the current agents' observed trajectories and HD maps as the input, when the local behavior data is unavailable. Inspired by recent development in knowledge distillation, we use our pre-trained LBA prediction framework as the teacher network during the training phase. This teacher network guides the LBF student network in inferring the features of the absent local behavior data. The intuition behind this design is that a traffic agent's movement at a particular location is confined to a limited number of possibilities. The teacher network essentially provides the ground truth of the movement pattern, making it plausible for the student network to learn the inference of the pattern given the current scene information.

LBA and LBF frameworks both have strong generalizability to be adopted by a wide range of existing trajectory forecasting methods. In Sec. \ref{sec:lba_two_cases} and Sec. \ref{sec:lbf_two_cases}, we showcase the implementation methodology on how to upgrade existing systems to the LBA/LBF framework respectively. We then implement and validate the two frameworks based on three state-of-the-art trajectory forecasting methods, P2T~\cite{deo2020trajectory}, LaneGCN~\cite{liang2020learning}, and DenseTNT~\cite{gu2021densetnt}.

In summary, this work has three major contributions to the literature:
\begin{itemize}
    
    
    \item We propose a Local-Behavior-Aware prediction framework. It enables most of the existing methods to incorporate the local behavior data, a new type of system input which contains the local historical information.
    
    \item We further introduce a Local-Behavior-Free prediction framework, which adopts a knowledge-distillation-based architecture to infer local behavioral information when it is temporarily unavailable.

    \item We conduct extensive experiments on published benchmarks (Argoverse~\cite{chang2019argoverse}, nuScenes~\cite{caesar2020nuscenes}), and validate that upgrading the SOTA methods to LBA/LBF frameworks consistently improves their performances by significant margins on various metrics. Especially, the LBA framework improves the SOTA methods on the nuScenes dataset by at least 14\% on the K=1 metrics.
\end{itemize}

\section{Related Work}\label{literature}
\subsection{Historical Behaviors in Trajectory Forecasting}
Historical behaviors are very helpful to trajectory forecasting since they reveal the motion pattern of agents. Several previous works have made significant progress in this direction by adopting memory-based designs. MANTRA~\cite{marchetti2020mantra} uses the observed past trajectories and the map as the key to query the possible hidden features of future trajectories. Similarly, Zhao et al.~\cite{zhao2021you} build an expert repository to retrieve the possible destinations by matching the current trajectory with the ones in the repository. MemoNet~\cite{xu2022remember} also considers memorizing the destinations of each agent and first applies the memory mechanism to multi-agent trajectory forecasting. Compared to those memory-based methods, our work: i) regards the historical behaviors as system inputs to benefit the task from the perspective of enriching scene information; ii) directly uses geometric coordinates to query related historical information which emphasizes the data locality and is more interpretable and robust.

\subsection{Scene Representation in Trajectory Forecasting}
To use a new type of scene information in systems requires fusing it with existing scene information sources. After reviewing how scene information is encoded in previous methods, we see two main types of scene representations: 1) rasterized-image-based representations~\cite{casas2018intentnet,hu2020collaborative}, 2) graph-based representations~\cite{gao2020vectornet,gilles2021gohome,liang2020learning,xu2022groupnet}. Rasterized-image-based representations render static HD maps and motion data into a birds' eye view (BEV) image, using various colors to represent different objects. Systems with this scene representation tend to use standard convolutional neural networks (CNNs) backbones to extract scene features~\cite{casas2018intentnet,deo2020trajectory,gilles2021home,hu2020collaborative,phan2020covernet}. These methods transform scenes into image coordinates systems. 

Graph-based scene representations become popular with the recent development of graph learning~\cite{scarselli2008graph} and attention mechanism~\cite{vaswani2017attention}. These methods build a graph that can be either directional or non-directional, and use techniques such as graph neural network (GNN)~\cite{scarselli2008graph} or attention-based operations~\cite{vaswani2017attention} to enable the interaction among map objects and agents. The nodes are the semantic map objects, e.g., lane segments and agents. The edges are defined by heuristics, which can be the spatial distances between the two nodes or the semantic labels of the two nodes. Systems using graph-based scene representations~\cite{gao2020vectornet,gilles2021gohome,gu2021densetnt,liang2020learning,zhao2020tnt} independently encode each map object to make the graph more informative. Graph-based scene representations have an outstanding information fusion capability and are substantially explored recently.

Other methods that do not strictly fall into the above two categories can share some properties with one or both of them. For example, TPCN~\cite{ye2021tpcn} uses point cloud representations for the scene and does not manually specify the interaction structures. Yet, by using PointNet++~\cite{qi2017pointnet++} to aggregate each point's neighborhood information, TPCN still technically defines local complete subgraphs for the scene where each point is a node connected with its neighbors.

To show the generalizability of the proposed frameworks, in this work we introduce an implementation methodology for upgrading forecasting systems that use either rasterized-image-based or graph-based scene representations to the LBA and LBF frameworks in Sec \ref{sec:lba_two_cases} and \ref{sec:lbf_two_cases} respectively.

\subsection{Knowledge Distillation}

Our LBF framework is inspired by the recent study of knowledge distillation (KD). Knowledge distillation is a technique that compresses a larger teacher network to a smaller student network by urging the student to mimic the teacher at the intermediate feature or output level~\cite{hinton2015distilling}. KD is widely used for various tasks, including object detection~\cite{chen2017learning,hao2019end}, semantic segmentation~\cite{he2019knowledge,liu2019structured} and tracking~\cite{liu2019teacher}. Its usage is still being explored. For example, researchers start to use KD for collaborative perception tasks~\cite{li2021learning}. In our work, the proposed LBF framework uses KD in the trajectory forecasting task. Compared to previous works that pay more attention to model compression, we seek to compress the volume of input data. We use the offline framework as the teacher network, which takes each agent's local behavior data as the input along with HD maps and the agents' observed trajectories, while the student network (i.e., the online framework) only uses the later two data modalities as the input without requiring local behavior data. Our experiments show that KD-based information compression significantly boosts the performance of trajectory forecasting.






\section{Formulation of Local Behavior}
\label{sec:formulation}
In trajectory forecasting, the observed trajectories of the agents previously passing through a location may help infer the future trajectory of an agent currently at the location. In this work, we collect such historical information and reformulate it to make it become one of the inputs for the trajectory forecasting system. We name this new type of scene information as the \textit{local behavior data}. In this section, we introduce its formulation and the methodology about how to retrieve such data from existing datasets.


Consider a trajectory forecasting dataset with $S$ scenes and the $p$th scene has $K_{p}$ agents. The observed trajectory and ground truth future trajectory of agent $i$ in scene $p$ are denoted respectively as $\mathbf{X}^{(i,p)}$ and $\mathbf{Y}^{(i,p)}$, where $\mathbf{X}^{(i,p)}\!\in\!\mathbb{R}^{\mathbf{T}^{-}\times 2}$ and $\mathbf{Y}^{(i,p)}\!\in\!\mathbb{R}^{\mathbf{T}^{+}\times 2}$. Each $\mathbf{X}^{(i,p)}$ or $\mathbf{Y}^{(i,p)}$ consists of two-dimensional coordinates at $\mathbf{T}^{\!-}$ or $\mathbf{T}^{\!+}$ timestamps. Note that the coordinate system used for the trajectories is the global coordinate system aligned with the global geometric map. In this work, we specifically denote two special items in $\mathbf{X}^{(i,p)}$. We use $\mathbf{X}_{1}^{(i,p)}\!\in\!\mathbb{R}^{2}$ to represent the location of agent $i$ in scene $s$ at the \textit{first} timestamp, i.e., the first observed location of the agent. Accordingly, we use $\mathbf{X}_{\mathbf{T}^-}^{(i,p)}\!\in\!\mathbb{R}^{2}$ to denote the agent location at timestamp $\mathbf{T}^-$, i.e., its current location. By gathering all the observed trajectories in this dataset, we build a behavior database $\mathcal{D}_\mathcal{B} = \{\mathbf{X}^{(i,p)}, p\in \{1,2,\cdots,S\},i\in \{1,2,\cdots, K_p\}\}_{i,p}$. We can query the local behavior from $\mathcal{D}_\mathcal{B}$; namely, the local behavior data of agent $i$ in scene $p$ is
\begin{equation}
    \mathcal{B}^{(i,p)}_\epsilon = \{\mathbf{X}^{(j,q)}| \left \|\mathbf{X}_{\mathbf{T}^-}^{(i,p)}-\mathbf{X}_1^{(j,q)}  \right \|_{2} < \epsilon ,\mathbf{X}^{(j,q)}\in \mathcal{D}_\mathcal{B}\},
\end{equation}
where $\epsilon$ is an adjustable hyper-parameter defining the radius of the neighboring area of a location. The~\emph{size} of $\mathcal{B}^{(i,p)}_\epsilon$ refers to the number of observed trajectories in $\mathcal{B}^{(i,p)}_\epsilon$. Fig. \ref{fig:teaser} (B) shows the steps to query local behavior data from $\mathcal{D}_\mathcal{B}$.



\begin{figure}[htbp!]
    \centering
    \includegraphics[width=1\textwidth]{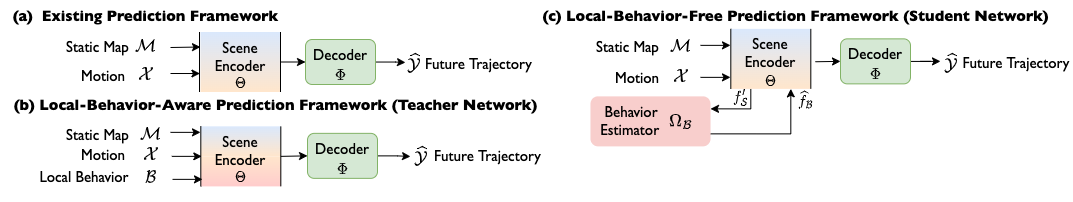}
    \caption{General pipelines of: (a) the baseline model that uses HD map data and observed trajectory data, which is typical for SOTA trajectory forecasting systems; (b) the LBA framework that takes local behavior data as an extra input; (c) the LBF framework that uses estimated local behavior features during the inference phase. (c) has fewer constraints in the use cases compared to (b).}
    \label{fig:pipelines}
\end{figure}

\section{Local-Behavior-Aware Prediction}\label{sec:generic-framework}
In real life, local behavior data has been widely collected by navigation apps and car companies. To use such data to benefit the trajectory forecasting task, we propose a Local-Behavior-Aware (LBA) prediction framework. In this section, we first demonstrate the generic pipeline of the LBA prediction framework and its major components. Then, we introduce the implementation strategy of the framework which especially emphasizes the representation of local behavior data and its corresponding scene encoder design. 


\subsection{Framework Pipeline}

Like a typical prediction framework pipeline adopted by previous works~\cite{gao2020vectornet,hu2020collaborative,phan2020covernet} (see Figure \ref{fig:pipelines} (a)), our LBA framework follows an encoder-decoder structure (see Figure \ref{fig:pipelines} (b)). The encoder $\Theta$ extracts features from multiple scene information sources. Then, the subsequent decoder $\Phi$ generates the predicted future trajectories $\widehat{\mathcal{Y}}$ based on the scene features.

The main difference between LBA and previous frameworks is the design of the encoder $\Theta$. Our scene encoder involves two steps: 1) scene modeling, which assembles data from all three sources to generate a comprehensive representation of the current scene, and 2) scene feature extraction, which extracts the high-dimensional features from the generated scene representation. The way to represent a scene would largely determine the inner structure of a scene encoder.





In the current literature, the two most frequently used scene representations are \textbf{graph}-based and \textbf{rasterized-image}-based representations. To show that most of the existing forecasting systems can be upgraded to fit the proposed LBA prediction framework, we will describe the implementation strategies for the graph-based and rasterized-image-based systems. 

\subsection{Implementation}\label{sec:lba_two_cases}

\subsubsection{Graph-based Systems.} 
The existing graph-based systems represent the whole scene information into a scene graph $G(V,E)$, where $V$ is the node set that contains both the map object node set $V_\mathcal{M}$ and the agent node sets $V_\mathcal{X}$, and $E$ is the edge set that reflects the internal interactions between nodes. For each map object node, its associated node attributes include the geometric coordinates, reflecting the static physical location of the map object. For each agent node, its associated node attributes include the two-dimensional coordinates across various time-stamps, reflecting the movement information of the agent. The characteristics of the graph-based representation are that i) it is compact and effective, leading to an efficient system; ii) it enables effective modeling of the interactions among objects (both map objects and agents) in the scene, which is crucial for understanding complicated dynamic relations in traffic scenarios. 
 
To implement the graph-based LBA prediction system, we emphasize the strategy of incorporating local behavior data into the scene graph and enabling the feature interaction between local behavior data and other nodes.
  



\textbf{\textit{Representation of Local Behavior Data.}} Given the $i$th agent in the $p$th scene, we can query its specific local behavior data $\mathcal{B}^{(i,p)}_\epsilon$ from the behavior database $\mathcal{D}_\mathcal{B}$. For each individual observed trajectory in  $\mathcal{B}^{(i,p)}_\epsilon$, we create a local behavior node. This step results in a local behavior node set $V_\mathcal{B}$, which has the same size as the local behavior data.

\begin{figure}[htbp!]
\centering
\begin{minipage}[t]{0.5\textwidth}
    \centering
    \includegraphics[width=1.\textwidth]{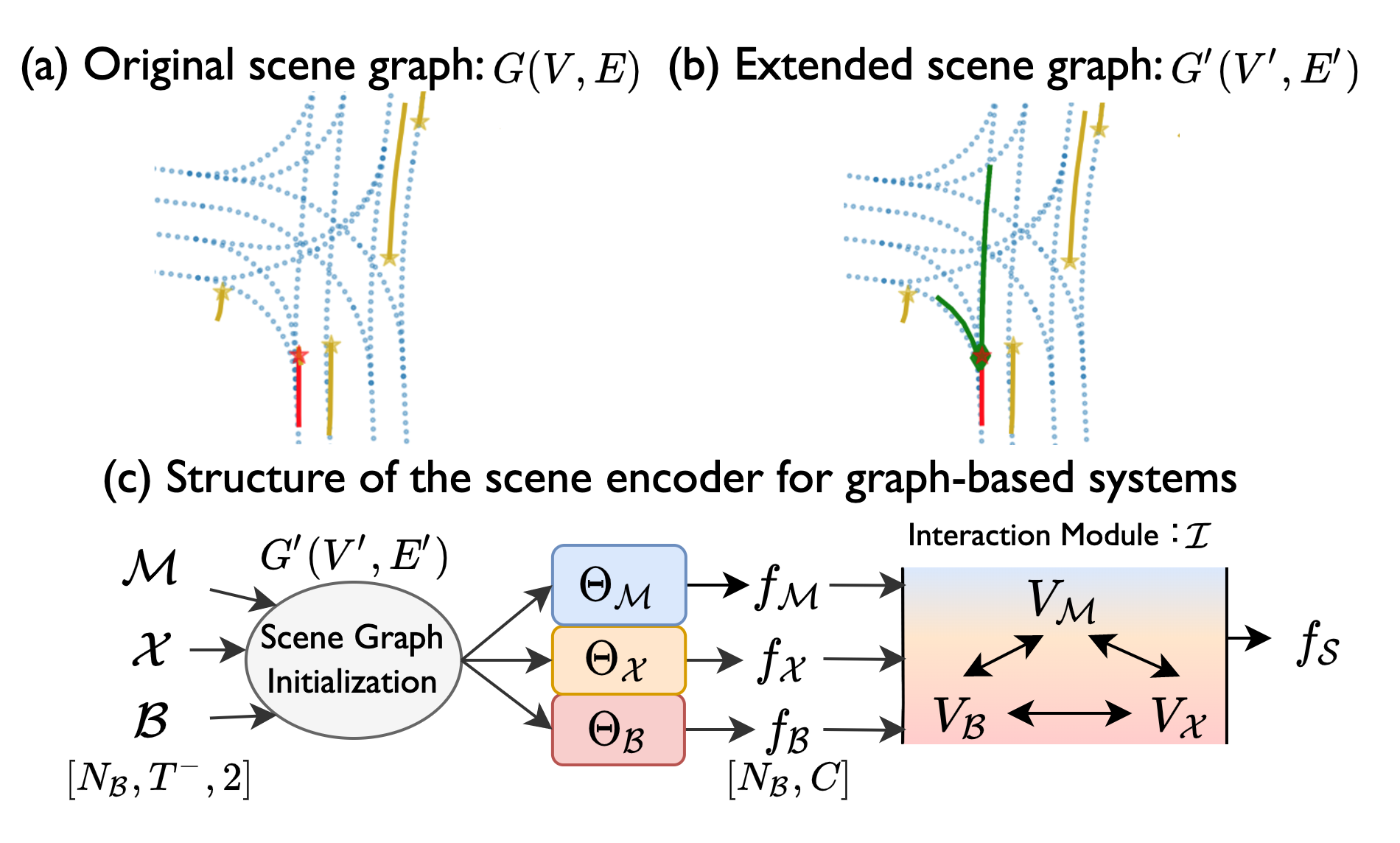}
    \caption{Scene encoder implementation for graph-based systems. In (c): i) $V_\mathcal{B},V_\mathcal{X},V_\mathcal{M} \subseteq V'$ are respectively the behavior node set, motion node set, and map node set; ii) $N_\mathcal{B}$ is the size of the local behavior data; iii) $C$ is the feature channel; iv) each trajectory in the local behavior data is encoded independently.}
    \label{fig:interaction-graph}
\end{minipage}
\hspace{2mm}
\begin{minipage}[t]{0.45\textwidth}
      \centering
    \includegraphics[width=1\textwidth]{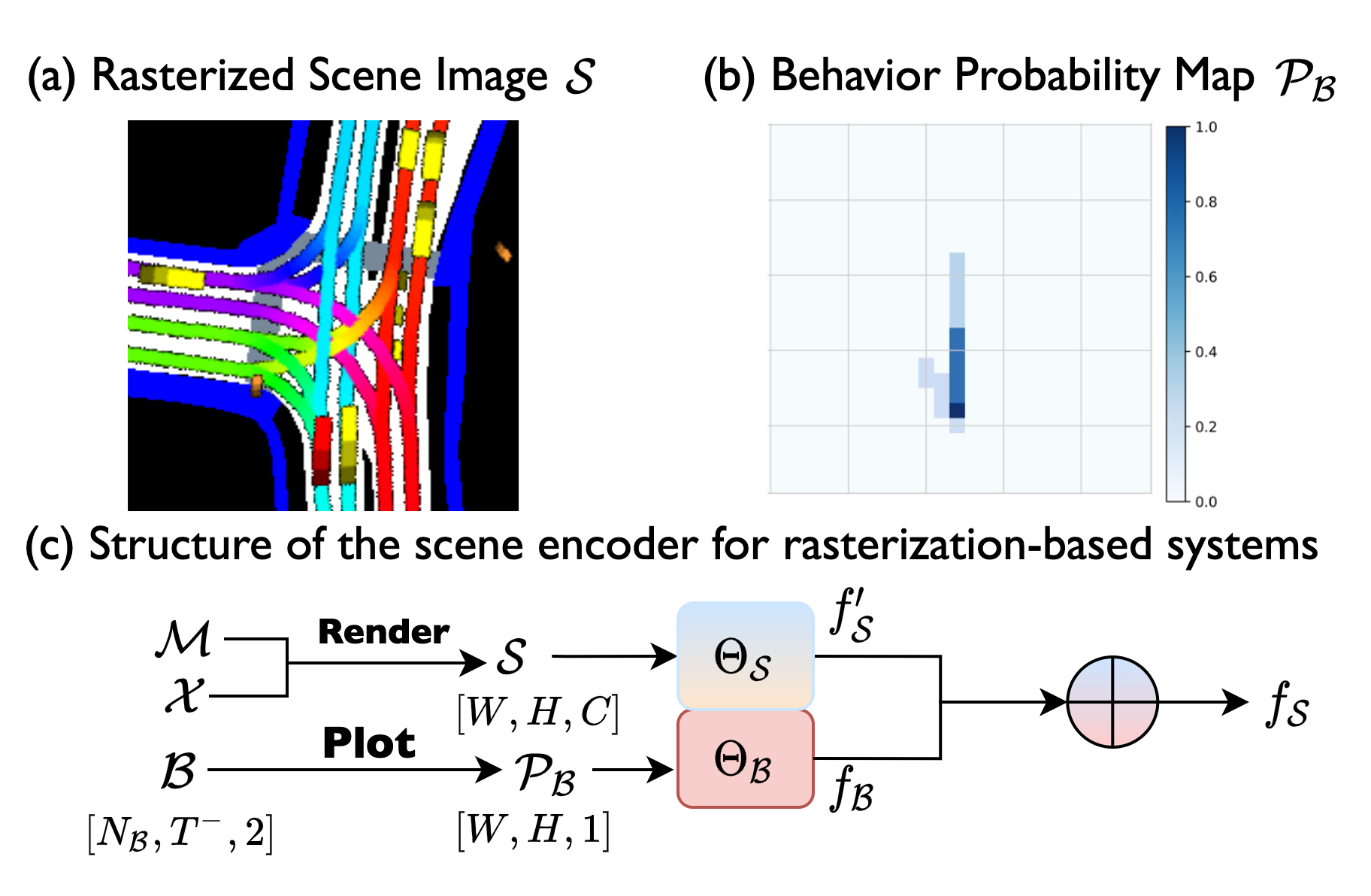}
    \caption{Scene encoder implementation for rasterized-image-based systems. (a) is generated with P2T's~\cite{deo2020trajectory} official code. In (c): features of the rasterized scene image and the local behavior are concatenated channel-wise to generate the final output features.}
    \label{fig:rasterized_image}
\end{minipage}
\end{figure}

\textbf{\textit{Scene Encoder.}} As demonstrated in Fig. \ref{fig:interaction-graph}, the scene encoder of the LBA graph-based systems includes the scene graph initialization, individual node feature extraction, and interactive node feature extraction. To initialize the local-behavior-aware scene graph $G'(V',E)$, we add the local behavior node set to the original graph $G(V,E)$, where the updated node set is $V' = V \cup V_\mathcal{B}$ and the updated edge set is $E' = E \cup \{(v_m,v_n)|v_m\in V',v_n\in V_\mathcal{B}\}_{m,n}$. With this graph, the local behavior data will participate in the feature interaction procedure (some methods may further update the edge set based on the node distance~\cite{li2021learning,zeng2021lanercnn,gu2021densetnt}). The output of the scene encoder $f_\mathcal{S}$ will also be local-behavior-aware. We use three feature encoders ($\Theta_\mathcal{M}$, $\Theta_\mathcal{X}$, $\Theta_\mathcal{B}$) to extract the features of map objects $f_\mathcal{M}$, agents' observed trajectories $f_\mathcal{X}$, and the local behavior data $f_\mathcal{B}$, respectively. Here each scene node will obtain its corresponding node features individually. To capture internal interactions, we use an interaction module $\mathcal{I}$, which is either GNN-based or attention-based, depending on the design of the original system. The interaction module aggregates information from all three scene components. 

The architecture of $\Theta_\mathcal{M}$, $\Theta_\mathcal{X}$ and $\mathcal{I}$ can remain unchanged from the original system structure for our implementation. As for $\Theta_\mathcal{B}$, since each independent behavior data is essentially an observed trajectory, we can directly adopt the structure of the trajectory encoder $\Theta_\mathcal{X}$ in the system as our behavior encoder structure. We can also use simple encoder structures, such as multi-layer perceptrons (MLPs), to keep the system light-weight. 

\subsubsection{Rasterized-image-based Systems.} 
Rasterized-image-based systems represent the whole scene information as a rasterized scene image (see an example in Fig. \ref{fig:rasterized_image}). The rasterized scene image reflects the HD map objects (junctions, lanes) and the agent trajectory information as BEV images with various colors according to their semantic labels. This representation method essentially transforms a global coordinate system into an image coordinate system, where each location on the map can be represented by a pixel coordinate. The main characteristic of rasterization-based representations is that they can leverage established CNN backbones, such as ResNet~\cite{he2016deep}, to extract image features. 

To implement the rasterized-image-based LBA prediction system, we need to render local behavior data to an image that has the same coordinate system with the original scene image, which ensures consistency and compatibility.


\textbf{\textit{Representation of Local Behavior Data.}} We seek to render local behavior data into a \emph{behavior probability map} $\mathcal{P}_{\mathcal{B}}$. It is an image whose pixel value reflects an agent's moving probability from the current pixel to another pixel in the image. For the $i$th agent at the $p$th scene, $\mathcal{P}^{(i,p)}_{\mathcal{B}}$ is formulated as a single channel image of size $(H,W,1)$ that shares the same coordinate system with the rasterized scene image, where $W,H$ are the width and the height of $\mathcal{P}_{\mathcal{B}}$. To generate such a behavior probability map, we initialize each pixel value in the image $\mathcal{P}^{(i,p)}_{\mathcal{B}}(x,y)$ to be 0. We then enumerate each trajectory $f_{\mathcal{X}} \in \mathcal{B}_{\epsilon}^{(i,p)}$ and add the corresponding information into $\mathcal{P}^{(i,p)}_{\mathcal{B}}(x,y)$ by adding 1 to the pixel value once each trajectory covers that pixel. In the end, we normalize $\mathcal{P}^{(i,p)}_{\mathcal{B}}$ by dividing every pixel by the maximum value of the pixels, specifically
\begin{equation}
    \mathcal{P}^{(i,p)}_{\mathcal{B}}(x,y) = \frac{\mathcal{P}^{(i,p)}_{\mathcal{B}}(x,y)}{\max(\mathcal{P}^{(i,p)}_{\mathcal{B}}(x,y))},
\end{equation}
where $0\leq x < W,0\leq y< H$. Fig.~\ref{fig:rasterized_image} (b) illustrates the local behavior data of the agent represented as a red rectangle in Fig.~\ref{fig:rasterized_image} (a). Since the configuration of the rasterized scene image and the behavior probability map are the same, this probability map indicates how likely, according to local behavior data, the agent at the current pixel will pass a certain pixel on the scene image.

\textbf{\textit{Scene Encoder.}} 
As shown in Fig.~\ref{fig:rasterized_image}, the scene encoder of the rasterized image-based LBA systems includes image rendering, image feature extraction, and concatenation-based aggregation. To achieve scene modeling, we render both the rasterized scene image and the behavior probability map. To comprehensively extract scene features, we use two separate encoders, $\Theta_\mathcal{S}$, which extracts features from the rasterized scene image $\mathcal{S}$, and $\Theta_\mathcal{B}$, which extracts features from the behavior probability map $\mathcal{P}_{\mathcal{B}}$. The architecture of $\Theta_\mathcal{S}$ can remain identical with the one in the original systems. To implement $\Theta_\mathcal{B}$, 
we can adopt any established images' feature extraction networks, such as ResNet~\cite{he2016deep}. Afterwards, the output of the two encoders will be concatenated channel-wise to build the output of the scene encoder, which is now local-behavior-aware.

\section{Local-Behavior-Free Prediction}\label{sec:data-free}
As mentioned in Sec \ref{sec:intro}, there will be scenarios where the local behavior data is yet to be gathered or insufficient. To handle such situations, we propose a Local-Behavior-Free (LBF) prediction framework based on knowledge distillation. 

The training of the LBF prediction framework follows a teacher-student structure. The teacher network is implemented using the LBA framework introduced in Sec \ref{sec:generic-framework}, which uses local behavior data as the third input; while the student network only takes static HD map information and the agents' observed trajectories as the input. The knowledge-distillation-based training strategy enhances the training of the LBF prediction framework by urging the student network to imitate the teacher network in the intermediate feature levels when processing the same data samples. 

The intuition behind this design is that given a specific location, the number of possible movement patterns of an traffic agent is limited. With the guidance from the teacher network that is trained on local behavior data, it is feasible for the student network to learn the reasoning of the movement pattern based on the static map objects and agents' observed trajectories.

\subsection{Framework Pipeline}
The LBF framework includes a teacher network and a student network. See Fig. \ref{fig:pipelines} (b), (c). The teacher network follows the LBA framework. For the student network, we remove the input stream of local behavior data in the LBA framework and add a behavior estimator $\Omega_\mathcal{B}$ to the pipeline. $\Omega_\mathcal{B}$ takes the intermediate scene features $f'_\mathcal{S}$ from the scene encoder $\Theta$ as the input, and outputs the estimated local behavior features $\widehat{f}_\mathcal{B}$. Note that $f'_\mathcal{S}$ is not involved with local behavioral information. Next, we let $\widehat{f}_\mathcal{B}$ join the scene encoder along with $f'_\mathcal{S}$ for the final feature generation. The scene encoder outputs the updated $f_\mathcal{S}$, which contains the estimated local behavioral information, to the decoder $\Phi$. The decoder then processes $f_\mathcal{S}$ and generates the predicted future trajectories $\widehat{\mathcal{Y}}$. The core step of the LBF student network is to link the behavior estimator and the scene encoder.

\subsection{Implementation}\label{sec:lbf_two_cases}
Like Sec \ref{sec:lba_two_cases}, this section introduces the implementation of the student networks in the LBF prediction framework based on two scene representations.




\subsubsection{Graph-based Systems.}
\begin{figure}[htbp!]
\centering
\begin{minipage}{0.56\textwidth}
    \centering
    \includegraphics[width=1\textwidth]{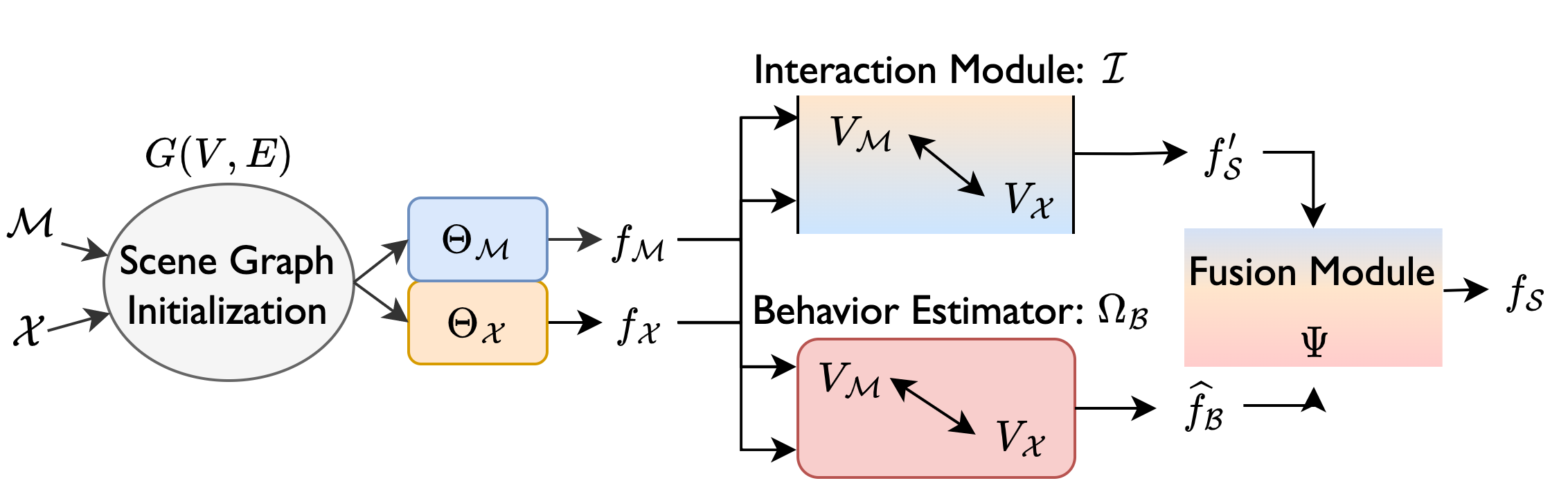}
    \caption{Local-Behavior-Free implementation for graph-based systems. During training, $\widehat{f}_\mathcal{B}$ and ${f}_\mathcal{S}$ can be supervised by the corresponding features from the teacher network.}
    \label{fig:student_graph}
\end{minipage}
\hspace{2mm}
\begin{minipage}{0.4\textwidth}
      \centering
    \includegraphics[width=1\textwidth]{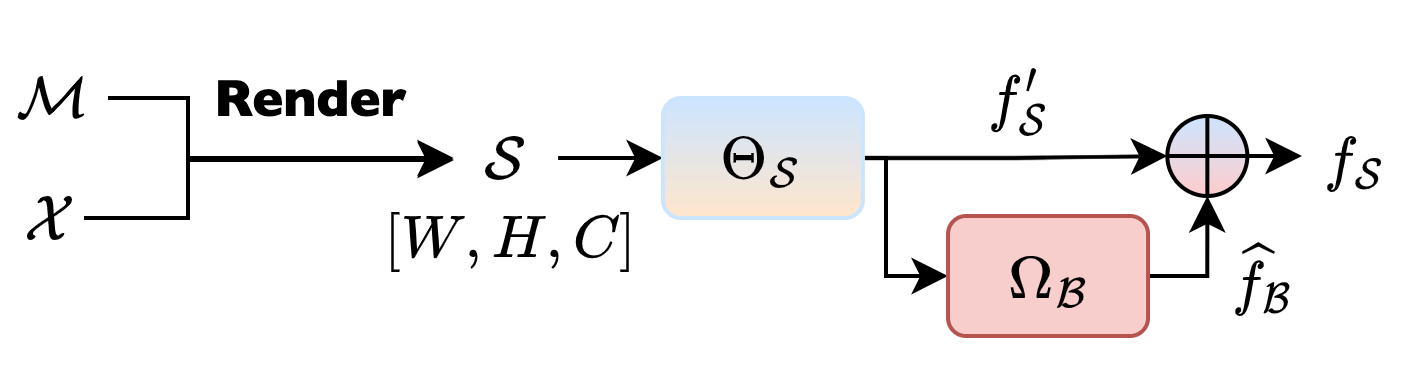}
    \caption{Local-Behavior-Free implementation for rasterization-based systems.}
    \label{fig:student_image}
\end{minipage}
\end{figure}

In the teacher network (the LBA prediction system), the local behavior data goes through an encoder to obtain behavior features $f_\mathcal{B}$. In the student network (the LBF version), we use a behavior estimator $\Omega_\mathcal{B}$ to estimate the behavior features even when the original local behavior data is not available. We implement the behavior estimator $\Omega_\mathcal{B}$ by a graph neural network. Its input includes the features of the map objects $f_\mathcal{M}$ and the features of the agents’ observed trajectories $f_\mathcal{X}$, which include all scene information in hand. It outputs the estimated behavior features $\widehat{f}_\mathcal{B}$. 
After $f_\mathcal{M}$ and $f_\mathcal{X}$ are interacted in the interaction module, we aggregate its output $\widehat{f}'_\mathcal{S}$ and the estimated behavior features $\widehat{f}_\mathcal{B}$ in a fusion module $\Psi$ to form the final scene feature $f_\mathcal{S}$. We implement $\Psi$ by an attention-based network. The pipeline of this procedure is shown in Fig. \ref{fig:student_graph}. During training, the estimated behavior features $\widehat{f}_\mathcal{B}$ in the student network can be supervised by the behavior features ${f_\mathcal{B}}$ in the teacher network through a knowledge-distillation loss.

\subsubsection{Rasterized-image-based Systems.}
In the teacher network (the LBA prediction system), the behavior probability map leads to behavior features $f_\mathcal{B}$. In the student network (the LBF version), we use a behavior estimator $\Omega_\mathcal{B}$ to estimate such behavior features even the behavior probability map is not available. We implement the behavior estimator $\Omega_\mathcal{B}$ by a CNN-based network. Its input is the output of the scene encoder $\Theta_\mathcal{S}$, which is the intermediate scene feature $f'_\mathcal{S}$. Its output is the estimated behavior features $\widehat{f}_\mathcal{B}$. We then concatenate $\widehat{f}_\mathcal{B}$ and $f'_\mathcal{S}$ to form the final scene feature $f_\mathcal{S}$. The pipeline of this procedure is shown in Fig. \ref{fig:student_image}. During training, the estimated behavior features $\widehat{f}_\mathcal{B}$ in the student network can be supervised by the behavior features ${f_\mathcal{B}}$ in the teacher network through a knowledge-distillation loss.

Note that both graph-based and rasterized-image-based LBF systems follow the same design rationale: estimating local behavior data based on the other known scene information. The difference between the two types of systems is in the fusion step. In the graph-based system, we use a trainable fusion module; and in the rasterized-image-based system, since the scene image and the behavior probability map share the same coordinate system, we simply concatenate them.

\subsection{Training with Knowledge Distillation}
To train such a teacher-student framework, we pre-train an LBA prediction network as the teacher network, and then use the intermediate features from the teacher network as the supervision signals to guide the training of the student network. We consider features from the teacher network that are leveraged to guide the student network training as $\mathbf{F}_t$ and the corresponding features from the student network as $\mathbf{F}_s$. Note that $\mathbf{F}_s$ may include but are not limited to the reconstructed local behavior features and the the scene encoder's outputs . 

The training loss thereafter contains the trajectory forecasting loss $\mathcal{L}_{pred}$, which is defined by the original system and identical with the training loss of the teacher network as well as a KD loss. In this work, we use $\ell_2$ loss as the KD loss and set $\lambda_{kd}$ as the adjustable weight of the KD loss. The overall loss function aggregates the loss over all the samples; that is,
\begin{equation}
    \mathcal{L} = \Sigma_{i}{ \left(\lambda_{kd}\left \|\mathbf{F}^i_{s}-\widehat{\mathbf{F}}^i_{t} \right \|_{2} + \mathcal{L}^{i}_{pred} \right)}.
\end{equation}

\section{Experiments}\label{experiment}


\subsection{Experimental Setup}\label{sec:dataset}

\textbf{Datasets.} We consider two widely used trajectory forecasting benchmarks, \emph{nuScenes}~\cite{caesar2020nuscenes} and \emph{Argoverse}~\cite{chang2019argoverse}. \emph{nuScenes} collected 1000 scenes. Each scene is 20s long with a sampling rate of 2 Hz. The instances for trajectory forecasting tasks were split into train, validation and test splits, which respectively entail 32186, 8560 and 9041 instances. Each agent in the instances had 2s' observed trajectories and the ground truth of 6s' future trajectories. \emph{Argoverse} collected over 30K scenarios. Each scenario is sampled at 10 Hz. The train/ val/ test splits had 205942/ 39472/ 78143 sequences respectively. Each agent in the scene had 2s' observed trajectories and the ground-truth of 3s' future trajectories. 

\textbf{Database Construction.}
To evaluate local-behavior-aware (LBA) framework without ground-truth data leakage, we only use the 2s' observed trajectories of all agents in each data split to build the behavior database $\mathcal{D_B}$ for each corresponding phase (e.g. testing phase only uses the test split). For the local-behavior-free (LBF) framework, we only use the observed trajectories from the training split during the training phase. Please see appendix for detailed information of behavior database construction.

\textbf{Metrics.} We adopt three widely used metrics for quantitative evaluation: minimum average displacement error among $K$ predictions (minADE$_{K}$), minimum final displacement error among $K$ predictions (minFDE$_{K}$) and the missing rate (MR). minADE evaluates the minimum average prediction error of all timestamps among the predicted $K$ trajectories; minFDE is the minimum error of the final position among the $K$ predictions; MR$_K$ is the ratio of agents whose minFDE$_{K}$ is larger than 2.0 meters.

\subsection{Implementation Details}

We pick three SOTA trajectory forecasting methods:  LaneGCN~\cite{liang2020learning},  DenseTNT~\cite{gu2021densetnt} and P2T~\cite{deo2020trajectory}, and adapt them to our behavior-aware framework. We use their official code packages as the implementation start code. The baseline performances in Table \ref{tab:nuscenes-eval} and \ref{tab:argo-eval} are reproduced by ourselves for variables controlling.

LaneGCN and DenseTNT are graph-based methods. We use stacked linear residual blocks mentioned in \cite{liang2020learning} as our behavior encoder $E_\mathcal{B}$ for both methods. In the LBF prediction, we use attention based architecture to implement the behavior estimator $\Omega_\mathcal{B}$ as well as the auxiliary fusion module $\Psi$.

P2T is a rasterization-based method. We use \emph{behavioral probability map} to represent the local behavior data, and ResNet~\cite{he2016deep} as the encoder backbone to extract features from the behavioral probability map. For the LBF prediction, we use three 1D Convolutional Layers to implement the behavior estimator $\Omega_\mathcal{B}$. 

To train the network, we adopt the hyper-parameter configuration from each method's official instructions. More implementation details are in the appendix. 

\subsection{Evaluation}
We evaluate LaneGCN and P2T on nuScenes (see the quantitative results in Table \ref{tab:nuscenes-eval}). For Argoverse, we evaluate LaneGCN and DenseTNT (see Table \ref{tab:argo-eval}). We also show the qualitative evaluation in Fig. \ref{fig:qual}. 

In all experiments, when upgraded to either the LBA or LBF framework, the baseline methods significantly improve in performance. Unconventionally, our proposed frameworks bring consistent improvements to various models (P2T, LaneGCN, DenseTNT) across \textbf{all} metrics. This is a substantive progress compared to previous methods, because SOTA methods~\cite{gilles2021gohome,gilles2021thomas,ye2021tpcn,zeng2021lanercnn} usually only show improvements in one or some metrics but not all metrics. Furthermore, on both datasets, the gains brought by local behavior data are consistently larger on $K=1$ metrics than on other metrics. This may result from the raise of average performance of the worst prediction among \textit{K} predictions, as local behavior data efficiently narrows down the solution search space (explained in Sec \ref{sec:intro}).

\begin{table}[htbp!]
\setlength\tabcolsep{1pt}
\caption{Evaluation results on nuScenes~\cite{caesar2020nuscenes} dataset test split.}

    \centering
\scalebox{0.74}{
    \begin{tabular}{c|c|ll|ll}

{Method}&
{Framework}&
{minADE}$_{1}$ & {minFDE}$_{1}$ & {minADE}$_{10}$ & {minFDE}$_{10}$ \\
\hline
\hline
\multirow{3}{*}{{P2T}~\cite{deo2020trajectory}}


&{Baseline} & 4.60	&	10.80	&	1.17	&	2.15\\

& {LBA}	
&	3.78 \textcolor{red}{$\downarrow 18$\%}
&	9.25 \textcolor{red}{$\downarrow 14$\%}	
&	1.08 \textcolor{red}{$\downarrow 8$\%}	
&	2.04 \textcolor{red}{$\downarrow 5$\%}\\

& {LBF}	
&4.04	 \textcolor{red}{$\downarrow 12\%$}	
&9.54	 \textcolor{red}{$\downarrow 12\%$}	
&1.15	 \textcolor{red}{$\downarrow 2\%$}	
&2.11	 \textcolor{red}{$\downarrow 2\%$}  \\

\hline
\multirow{3}{*}{{LaneGCN}~\cite{liang2020learning}}&
{Baseline} &	6.17	&	12.34	&	1.82	&	2.98\\
& {LBA}&	2.72 \textcolor{red}{$\downarrow 56$\%}
&	6.78 \textcolor{red}{$\downarrow 45$\%}
&	0.95 \textcolor{red}{$\downarrow 48$\%}
&	1.85 \textcolor{red}{$\downarrow 38$\%}   \\
& {LBF}
&	5.58 \textcolor{red}{$\downarrow 10$\%}
&	11.47	\textcolor{red}{$\downarrow 7$\%}
&	1.67	\textcolor{red}{$\downarrow 8$\%}
&	2.66 \textcolor{red}{$\downarrow 11$\%}  \\
%

    \end{tabular}}
    
    \label{tab:nuscenes-eval}
\end{table}

Interestingly, the prediction performance of the LBF framework occasionally surpasses that of the LBA framework, even though the LBF framework lacks local behavior data input. Table \ref{tab:argo-eval} shows an example of this case, with the LaneGCN results on Argoverse. Our educated conjecture is that sometimes the LBF framework enjoys more data representativeness thanks to the reconstructed behavioral features; whereas, in the meantime, the LBA framework may be using pre-gathered local behavior data of a small size in its testing phase.

Besides the comparison with the baselines, we also show the comparison with the other published works on the benchmarks; see Tables \ref{tab:nusc-benchmark} and \ref{tab:argo-benchmark}.


\begin{table}[htbp!]
\setlength\tabcolsep{1.5pt}

    \centering
    \caption{Evaluation results on Argoverse~\cite{chang2019argoverse} dataset test split.}
    \scalebox{0.74}{
    \begin{tabular}{c|c|ll|ll}

{Method}&
{Framework}&
{minADE}$_{1}$ & {minFDE}$_{1}$ & {minADE}$_{6}$ & {minFDE}$_{6}$ \\
\hline
\hline 
\multirow{3}{*}{{LaneGCN}~\cite{liang2020learning}}
&{Baseline}   &	1.74	&	3.89	&	0.87	&	1.37 \\
&{LBA}	&	1.64 \textcolor{red}{$\downarrow 6\%$}	&	3.61 \textcolor{red}{$\downarrow 7\%$}
&	0.84 \textcolor{red}{$\downarrow 3\%$}	&	1.30 \textcolor{red}{$\downarrow 5\%$}\\
&{LBF}         &	1.61 \textcolor{red}{$\downarrow 7\%$}	&	3.54 \textcolor{red}{$\downarrow 9\%$}
&	0.85	\textcolor{red}{$\downarrow 2\%$}&	1.31 \textcolor{red}{$\downarrow 4\%$}\\
\hline
\multirow{3}{*}{{DenseTNT}~\cite{gu2021densetnt}}
&{Baseline}   &	1.70	&	3.72	&	0.90	&	1.33	\\
&{LBA}	&	1.65 \textcolor{red}{$\downarrow 3\%$}	&	3.57	\textcolor{red}{$\downarrow 4\%$}&	0.88 \textcolor{red}{$\downarrow 2\%$}	&	1.26 \textcolor{red}{$\downarrow 5\%$}	\\

&{LBF}    &1.67	\textcolor{red}{$\downarrow 2$\%}	&
3.63	\textcolor{red}{$\downarrow 2$\%}	&
0.89	\textcolor{red}{$\downarrow 2$\%}	&
1.29	\textcolor{red}{$\downarrow 3$\%}	
\\

    \end{tabular}}
    
    \label{tab:argo-eval}
\end{table}

\begin{table}[htbp!]
\centering
\begin{minipage}{0.5\textwidth}
    \centering
    \caption{nuScenes benchmark comparison.}
    \scalebox{0.74}{
    \begin{tabular}{c|c|c|c|c}
{Method}      &   {minADE}$_{10}$   &   {MR}${_5}$ & {MR}${_{10}}$  &   {minFDE}$_{1}$    \\
\hline
\hline
{P2T}\cite{deo2020trajectory}    &   1.16    &   64\%    &   46\%    &   10.50       \\
{MHA-JAM}\cite{messaoud2020trajectory}       &   1.24    &   59\%    &   46\%    &   8.57    \\
{SGNet}\cite{wang2021stepwise}       &   1.40    &   67\%    &   52\%    &   9.25    \\
{Trajectron++}\cite{salzmann2020trajectron++}     &   1.51    &   70\%    &   57\%    &   9.52    \\
{M-SCOUT}\cite{carrasco2021scout}   &   1.92    &   78\%    &   78\%    &   9.29    \\
{GOHOME}\cite{gilles2021gohome}  &   1.15    &   57\%    &   47\%    &   6.99    \\
\hline
\hline
{P2T-LBA}   &   1.08    &   57\%    &   41\%    &   9.25        \\
{P2T-LBF}   &   1.15    &   61\%    &   46\%    &   9.37        \\
{LaneGCN-LBA}   &   0.95    &   49\%    &   36\%    &   6.78        \\
{LaneGCN-LBF}    &   1.67    &   75\%    &   68\%    &   11.47       \\

    \end{tabular}}
    
    \label{tab:nusc-benchmark}
\end{minipage}
\begin{minipage}{0.44\textwidth}
      \centering
      \caption{Argoverse benchmark comparison.}
    \scalebox{0.74}{
    \begin{tabular}{c|c|c|c}

{Method}  &   {minADE}$_{1}$        &   {minADE}$_{6}$    &   {Brier-FDE}$_{6}$ \\
\hline
\hline
{LaneRCNN}\cite{zeng2021lanercnn} &   1.69         &   0.90        &   2.15    \\
{DenseTNT}\cite{gu2021densetnt}   &   1.68          &   0.88      &   1.98    \\
{GOHOME}\cite{gilles2021gohome}   &   1.69          &   0.94       &   1.98    \\
{MMTransformer}\cite{liu2021multimodal}   &   1.77           &   0.84       &   2.03    \\
{LaneGCN}\cite{liang2020learning} &   1.71          &   0.87       &   2.06    \\
\hline
\hline
{LaneGCN-LBA}   &   1.64    &  0.84   &   2.00  \\
{LaneGCN-LBF} &   1.61      &   0.85       &   2.00    \\
{DenseTNT-LBA} &  1.65          &   0.88    &   1.93     \\
{DenseTNT-LBF} & 1.67     &   0.89      &   1.96    \\

    \end{tabular}}
    
    \label{tab:argo-benchmark}
\end{minipage}
\end{table}

\begin{minipage}{0.56\textwidth}
    
    \includegraphics[width=1.005\textwidth]{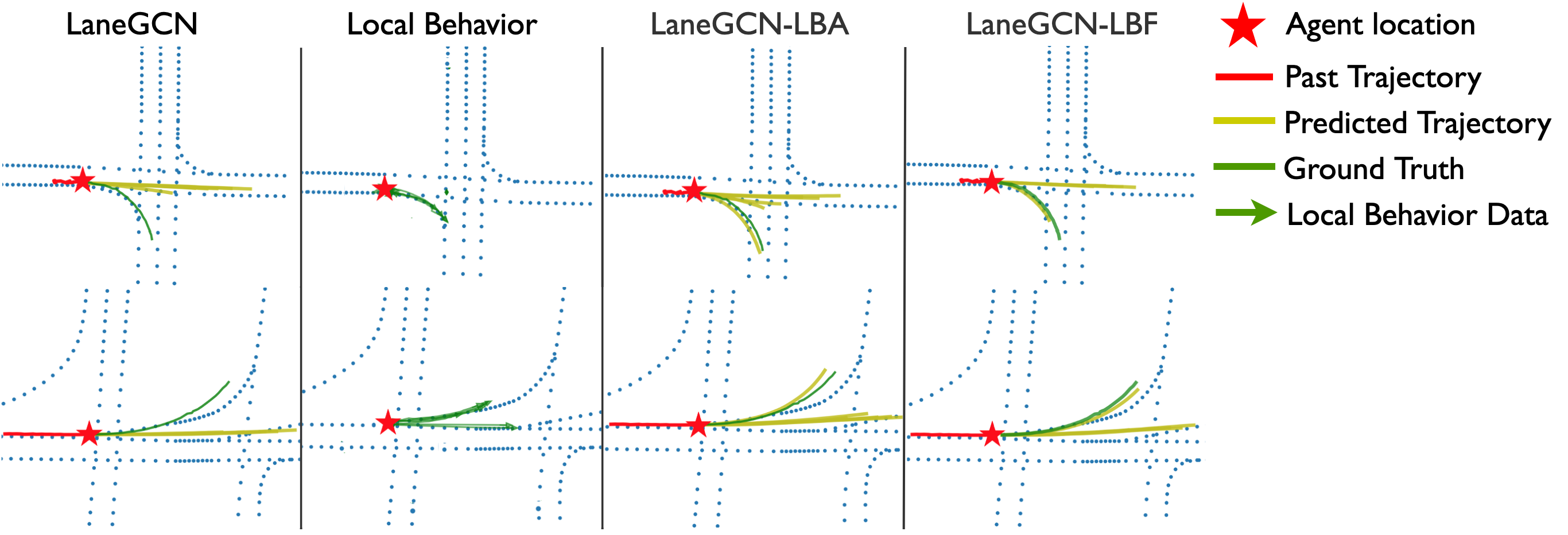}
  
    \captionof{figure}{LaneGCN qualitative results on\\ Argoverse val split. }
    \label{fig:qual}
\end{minipage}
\begin{minipage}{0.4\textwidth}

    \centering
    \captionof{table}{minFDE$_1$ of LaneGCN on Argoverse val split among agents having different sizes of local behavior data.}
    \scalebox{0.74}{
    \begin{tabular}{c|c||c|c}

{Size} &  {\% of agents} & {Baseline} & {LBA}\\
\hline
\hline

$\left[0,4\right)$ & 24\%&3.33 & 3.36 \textcolor{green}{$\uparrow1\%$}\\
$\left[4,8\right)$ & 14\%& 3.25 & 3.19 \textcolor{red}{$\downarrow2\%$} \\
$\left[8,12\right)$ & 10\%& 3.22 & 3.13 \textcolor{red}{$\downarrow3\%$}\\
$\left[12,16\right)$ & 9\%& 3.00 & 2.87 \textcolor{red}{$\downarrow4\%$}\\
$\left[16,\infty\right)$ & 43\%& 2.66 & 2.60 \textcolor{red}{$\downarrow2\%$} \\
\end{tabular}}

\label{table:size}
\end{minipage}

  

%

%


\subsection{Ablation Study}

We conduct the ablation study on the hyper-parameters regarding local behavior data generation and our prediction frameworks.

\textbf{Local Behavior Data Size.} Regarding the relationship between prediction performance and the local behavior data size of each agent, Table \ref{table:size} shows: when there is few local behavior data, prediction is not as accurate as the baseline; but, even a small size of local behavior data can improve the prediction. 

\textbf{Local Range $\epsilon$.} The adjustable parameter $\epsilon$ defines the local radius (see Sec \ref{sec:formulation}). See Table \ref{tab:ablation} for the impact of $\epsilon$ on LaneGCN performance with Argoverse test split. When $\epsilon$ increases, the performance of the LBA framework first go up but later go down. It matches our intuition that a large $\epsilon$ will introduce distractions. The LBF framework, however, shows less confusion brought by the pre-trained LBA network, demonstrating stronger robustness to $\epsilon$.


\textbf{Knowledge Distillation Parameters.} We study the impact of knowledge-distillation-related parameters, i.e., the loss weight $\lambda_{kd}$ and the number of times that intermediate features are involved in the KD loss (denoted as KD times). See the results in Table \ref{tab:weight_time}. Detailed information about the choices of intermediate features is in the appendix. We see that firstly, the knowledge distillation structure does help the framework infer the impact of local behavior data. The framework, across all parameter settings in Table \ref{tab:weight_time}, outperforms the baseline method (in Table \ref{tab:argo-eval}, row 1) and the model without KD loss (in Table \ref{tab:weight_time}, row 1). Secondly, the comparatively similar results across all the settings show that the LBF framework is relatively robust to the KD hyper-parameters.


\begin{table}[htbp!]

\centering
\begin{minipage}[t]{0.5\textwidth}
    \centering
    \caption{Impact of the behavioral data's local range. A larger $\epsilon$ brings a larger size of local behavior data, but when $\epsilon$ is too large, it can also introduce confusion due to the loss of the data locality. }
    \scalebox{0.74}{
        \begin{tabular}{c|c|cc}

{Framework}&
{$\epsilon$}&

{minADE}$_{1}$ & {minFDE}$_{1}$ \\
\hline
\hline

\multirow{3}{*}{{LBA}}
&0.5m   &   1.64  &   3.61  \\
&1.0m   &   1.62  &   3.58   \\
&1.5m   &   1.71  &   3.86  \\
     
\hline

\multirow{3}{*}{{LBF}} 
 &0.5m&    1.61 &   3.54 \\
 &1.0m &   1.62  &   3.57 \\
 &1.5m &   1.63   &   3.59  \\

    \end{tabular}}
    
    \label{tab:ablation}
\end{minipage}
\begin{minipage}[t]{0.5\textwidth}
\end{minipage}
\begin{minipage}[t]{0.35\textwidth}
      \centering
      \caption{Impact of the number of times that KD supervision gets applied in the training phase and the weight of the KD loss. }
    
    \scalebox{0.74}{
    \begin{tabular}{c|c|cc}

KD times&
{$\lambda_{kd}$}&
{minADE}$_{1}$ & {minFDE}$_{1}$ \\
\hline
\hline
N/A   &   0   &   1.68    &   3.72   \\
2   &   1   &   1.62  &   3.57  \\
2   &   1.5 &   \textbf{1.61} &   \textbf{3.54} \\
2   &   2   &   1.62  &   3.56  \\
1   &   1.5 &   1.64  &   3.61  \\
3   &   1.5 &   1.61  &   3.56 \\

    \end{tabular}}
    
    \label{tab:weight_time}
\end{minipage}
\end{table}


\section{Conclusion}
\label{conclusion}
In this work, we re-introduce the local historical trajectories as a new type of data input to the trajectory forecasting task, referred as \textit{local behavior data}. To adapt to this new data input and fully exploit its value, we propose a behavior-\textit{aware} framework and a behavior-\textit{free} framework for trajectory forecasting. The behavior-free framework, especially, adopts a knowledge-distillation architecture to estimate the impact of local behavior data. Extensive experiments on published benchmarks validate that the proposed frameworks significantly improve the performances of SOTA methods on prevalent metrics. 


\textbf{Limitations.} Local historical information reveals local motion patterns with high fidelity, but there are always outliers. For use cases of great safety concerns (e.g. autonomous driving), historical data may provide good reference but should not be the only reference. Also, the motion patterns of a certain location can vary over time. To optimize the benefits of the LBA and LBF framework, future research should explore the historical data gathering strategies.

\paragraph{Acknowledgement:} National Natural Science Foundation of China under Grant 62171276, the Science and Technology Commission of Shanghai Municipal under Grant 21511100900, CCF-DiDi GAIA Research Collaboration Plan 202112 and CALT 2021-01.


\par\vfill\par

\clearpage
%
%
\bibliographystyle{splncs04}
\bibliography{egbib}
\newpage

\appendix

\title{Appendix --- Aware of the History: Trajectory Forecasting with the Local Behavior Data} 
\titlerunning{Aware of the History: Trajectory Forecasting with the Local Behavior Data}
%
\author{Yiqi Zhong\inst{1}\orcidID{0000-0002-0928-8018} \and
Zhenyang Ni\inst{2}\orcidID{0000-0001-7134-620X} \and 
Siheng Chen\inst{2}\orcidID{0000-0001-6199-529X}\and 
Ulrich Neumann\inst{1}\orcidID{0000-0001-8977-7112}}
\authorrunning{Y. Zhong et al.}
%
\institute{University of Southern California, Los Angeles, CA 90089, USA \email{\{yiqizhon,uneumann\}@usc.edu}
\and
Shanghai Jiao Tong University, Shanghai, China\\
\email{\{0107nzy,sihengc\}@sjtu.edu.cn}}
\maketitle
\section{Dataset Construction}
In this work, we use two widely used autonomous driving benchmark: \emph{nuScenes}\cite{caesar2020nuscenes} and \emph{Argoverse}\cite{chang2019argoverse}. Here we describe how we derive local behavior data from each dataset for the experiments use. 

\subsection{nuScenes}
nuScenes collects scene from four different places from Singapore and Boston, USA. The four places are labeled as \emph{singapore-onenorh}, \emph{boston-seaport}, \emph{singapore-queenstown} and \emph{singapore-hollandvillage} by the dataset. For each data split, we collect all the 2-second observed trajectories in the four places separately to build the behavior database $\mathcal{D}_\mathcal{B}$. Since the sample rate of nuScenes is 2Hz, each 2-second observed trajectory suppose to have two-dimensional coordinates at 5 timestamps. However, in the dataset, there are missing timestamps for some observed trajectories. To make the learning procedure more stable, we only pick the observed trajectories those have no missing data. Meanwhile, we also filter out the static trajectories (whose speed is lower than 2m/s). Table \ref{tab:nusc_database_stat} shows the size of the behavior database for each data split in nuScenes dataset. 

\begin{table}[tbhp!]
    \centering
        \caption{The number of historical observed trajectories contained in the behavior database for each location in \emph{nuScenes} dataset. We build the database for each split separately to simulate the real-world scenario.}
    \begin{tabular}{c|c|c|c}
Location Label    & train    & val   & test  \\
\hline
singapore-onenorth	&	92893	&	54878	&	50570	\\
boston-seaport	&	708527	&	226813	&	163812	\\
singapore-queenstown	&	59702	&	3696	&	26568	\\
singapore-hollandvillage	&	92924	&	N/A	&	7340	\\
         
    \end{tabular}

    \label{tab:nusc_database_stat}
\end{table}
\subsection{Argoverse}
Argoverse collects data from Pittsburgh and Miami. We build the behavior $\mathcal{D}_\mathcal{B}$ for each city of each data split using the all 2-second observed trajectories correspondingly. In argoverse, since the sample rate is 10Hz, for each 2-second observed trajectory, there are 20 timestamps geometric position data. Similar to what we do for nuScenes, we only pick the observed trajectory that has no unavailable timestamp and whose average speed is larger than 2m/s. The statistic of the behavior database size is shown in Table \ref{tab:argo_database_stat}

\begin{table}[tbhp!]
    \centering
    \caption{The number of historical observed trajectories contained in the behavior database for each location in \emph{Argoverse} dataset.}
    \begin{tabular}{c|c|c|c}
Location    & train    & val   & test  \\
\hline

Miami	&	1532574	&	281809	&	538004	\\
Pittsburg	&	980827	&	162470	&	583311	\\
    \end{tabular}
    
    \label{tab:argo_database_stat}
\end{table}

\section{Extensive Discussion}
\subsection{Practicability.} When introducing a new data source to trajectory forecasting, researchers should evaluate its accessibility in real-world applications. Our experiments use a collection of available agents' observed trajectories in each dataset to build the behavior database $\mathcal{D}_\mathcal{B}$. The datasets used in this work both contain the \emph{global coordinates} for each recorded trajectory. This property of the datasets enables us to mount the trajectories to the physical location on the global map, which is fundamental to the concept of \textit{local behavior data}. Judging our framework in real-world applications, the global coordinates of the map and the agents are relatively easy to retrieve through techniques such as the Global Positioning System (GPS). Once the global coordinates are retrieved, gathering local behavior data in the real world becomes feasible. 

Furthermore, we argue that in the real-world application, the benefit brought by the local behavior data will be even more significant, since the behavior data gathering procedure in can last longer time and collect more data. Despite the improvement brought by the local behavior data in our experiments on the published datasets, in the Argoverse training split, there are still 5\% of agents have no available local behavior data.  

\subsection{Local Behavior Information Visualization}

\begin{figure}
    \centering
    \includegraphics[width=0.5\textwidth]{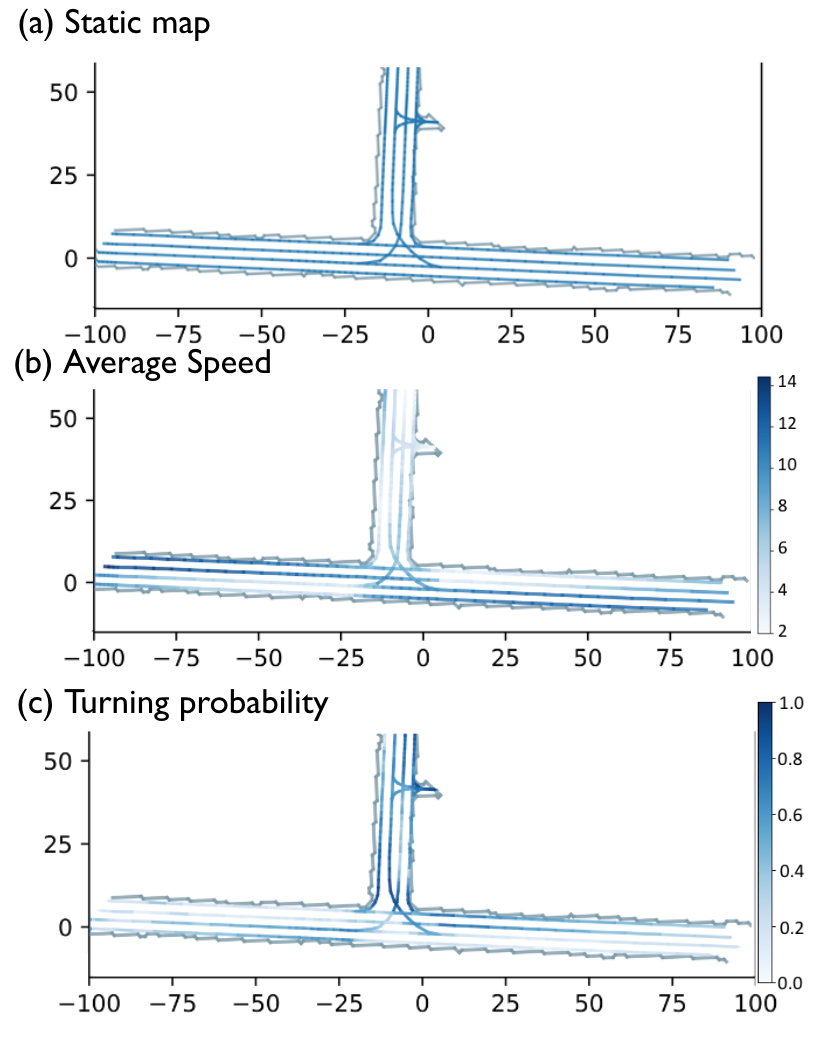}
    \caption{Visualization of the local behavior data for each lane segment.}
    \label{fig:behavior}
\end{figure}

In this paper, we exploit the local behavior data from the perspective of agent current locations. To more clearly visualize the information hidden in the local behavior data, here we show visualizations of the local behavior data for every lane segment on the map, see Figure \ref{fig:behavior}. 

In the figure, we focus on the visualization of two types of the information, one is the average speed of the trajectories collected from each lane segment and the other is the ratio of the trajectories that show turning actions (including turning and the lane changing). From the visualization we can see besides the location-specific information that is hidden in the local behavior data, such as the speed limit, there are more interesting information that worth further exploit. Those information includes some implicit rules of human driving behaviors that are hard to be learnt by limited number of observed trajectories, such as the turning intentions are much higher for the agents on the lanes that have intersections ahead even if the lane does not directly lead to a right-turn lane or left-turn lane. 

\section{Implementation}
We pick three SOTA methods for implementation. The implemented works in our paper are all based on existing official code packages. We \textbf{only modify the scene encoder related modules while remain the other parts exactly the same} as the baseline methods, including the decoders and the encoders for HD maps and motion data. For the LBF training, the network parameters are randomly initialized by pytorch, i.e. we did not use LBA teacher network as the initialize ion.
\subsection{LaneGCN\cite{liang2020learning}}
\textbf{Implementation details: }We upgrade LaneGCN to Local-Behavior-Aware and Local-Behavior-Free framework based on their official code (\hyperlink{https://github.com/uber-research/LaneGCN}{https://github.com/uber-research/LaneGCN}). During the training, on both nuScenes and Argoverse dataset, we use 2 NVIDIA GeForce RTX 3090 GPUs with the batch size of 32. The initial learning rate is 1e-3 and will decay to 1e-4 at epoch 32. The total training epoch number is 36. We replicate the exact training scheduler of the LaneGCN listed in the readme file of their official code package. We use $\epsilon = 0.5$ and $\lambda_{kd} = 1.5$ as the default setting.

\textbf{Dataset Preprocessing: }For the argoverse dataset, we use their official code to generate the preprocessed data, including the lane graph and the motion data. For the nuScenes dataset, we follow the instruction in their paper\cite{liang2020learning}, generating the lane graph using the lane information from nuScenes dataset. In the nuScenes, besides the lane information, there are also other map objects like sidewalks. Because in the LaneGCN paper, there is no indication for how to process those map objects, we decide to ignore them during our data preprocessing. It may explain that why the baseline performance of the LaneGCN on nuScenes (See Table  1 in the paper) is not as good as it shows on Argoverse. It also explains by the boost brought by the local behavior data is significant even compared to other methods: local behavior data provide extra complementary information to the system.

\begin{figure}[tbh!]
    \centering
    \includegraphics[width=0.8\textwidth]{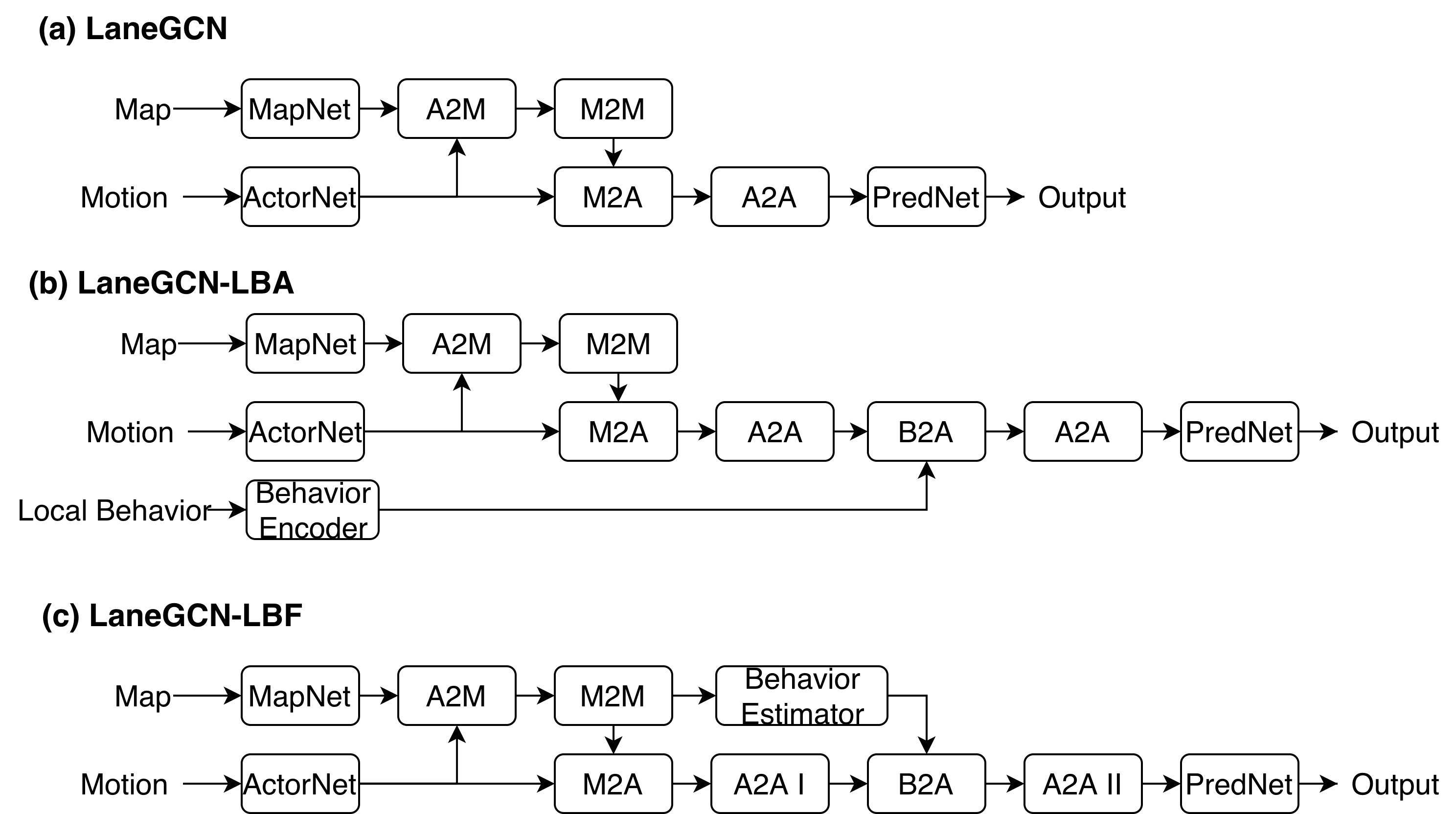}
    \caption{LaneGCN Network Architecture}
    \label{fig:lanegcn}
\end{figure}

\textbf{Network Architecture: }
We show the architecture of the implementation on LaneGCN based framework in Figure \ref{fig:lanegcn}.

In the network architecture drawn in the figure, B2A is the auxiliary fusion module for LBF framework. When applied knowledge distillation loss, we use the output of B2A, output of A2A II from the LBF as $\mathbf{F}_s$ and the output of B2A and the output of A2A as $\mathbf{F}_t$. In the Table 4 of the paper, to do the abalation study, we pick an intermediate layer inside the PredNet as the third supervised feature for comparison.

\begin{figure}[tbh!]
    \centering
    \includegraphics[width=0.7\textwidth]{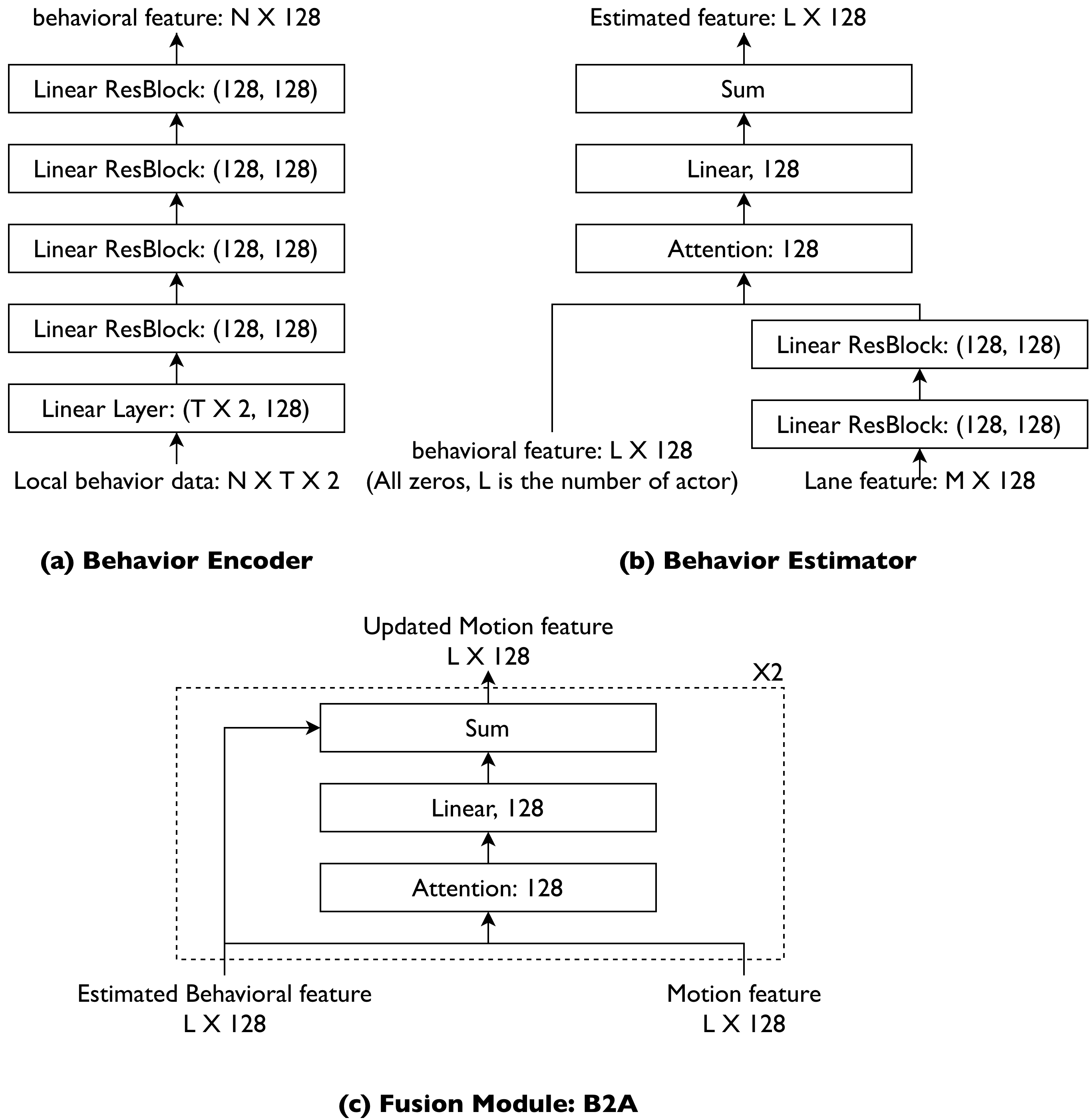}
    \caption{Innter structure of important module in LaneGCN implementation}
    \label{fig:inner-lanegcn}
\end{figure}

Detailed inner structure of behavior encoder, behavior estimator and auxiliary fusion module B2A are in Figure \ref{fig:inner-lanegcn} 
\subsection{DenseTNT\cite{gu2021densetnt}}
\textbf{Implementation Details. }In the experiment of DenseTNT, we adopt their official code package from \hyperlink{https://github.com/Tsinghua-MARS-Lab/DenseTNT}{https://github.com/Tsinghua-MARS-Lab/DenseTNT}. We use $\epsilon = 0.5$ and $\lambda_{kd} = 1.5$ as the default setting. We use 8 NVIDIA GeForce RTX 3090 GPUs with a batch size of 64 for training. Different from their paper which claims a two-stage training strategy, in their official code, they train the network in a end-to-end style. We adopt the training strategy in the official code and set the learning rate with an initial value of 0.001 decays to 30\% every 5 epochs. The total training epoch number is 30. The hidden size of the feature vectors is set to 128. The head number of our goal set predictor is 12. No data augmentation is used.

\textbf{Dataset Preprocessing: } We directly use their preprocess code to process the Argoverse dataset.

\begin{figure}[tbh!]
    \centering
    \includegraphics[width=.8\textwidth]{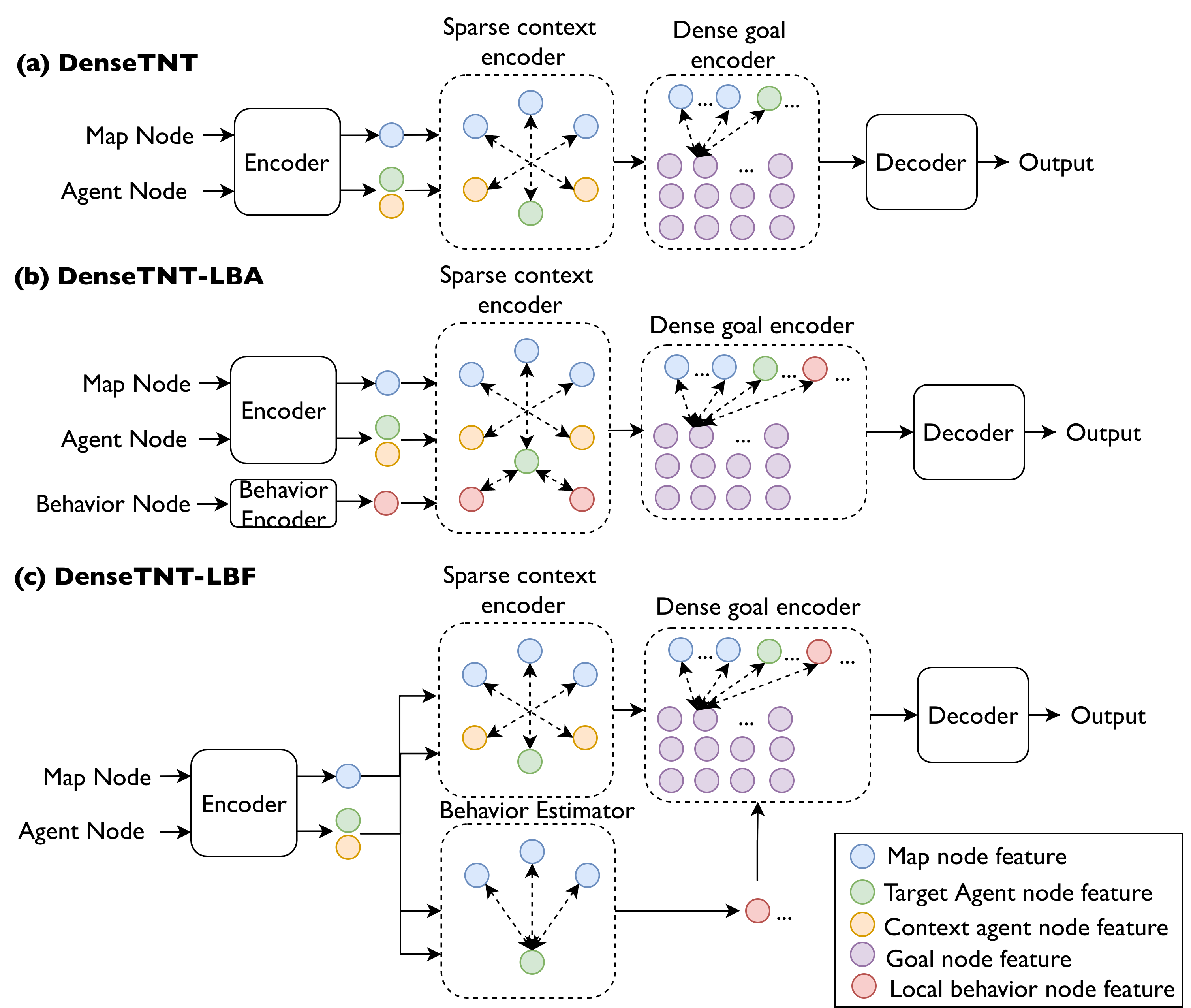}
    \caption{DenseTNT Implementation Architecture}
    \label{fig:densetnt}
\end{figure}

\textbf{Network Architecture: } We show the implemented architecture of DenseTNT in Figure \ref{fig:densetnt}. The inner architecture of behavior estimator and the behavior encoder are identical to the ones in the LaneGCN, which is shown in Figure \ref{fig:inner-lanegcn}. In the training of LBF framework ,$\mathbf{F}_s$ includes the output of Behavior estimator and the output of Dense goal encoder while $\mathbf{F}_t$ contains the target goal features of the Sparse context encoder and the output of the Dense goal encoder correspondingly.

\subsection{P2T\cite{deo2020trajectory}}
\textbf{Implementation Details: }The implementation of P2T is based on their published code package \hyperlink{https://github.com/nachiket92/P2T}{https://github.com/nachiket92/P2T}. For the baseline result, we directly use their released pre-trained model which is included in their code package. We use one NVIDIA GeForce 1080 Ti GPU for training and the batch size is 32. P2T trains the network in a three-stage style. It first trains a reward network for 25 epochs with the learning rate as 1e-4 and then trains a coarse trajectory predictor for 100 epochs whose learning rate is 1e-3. Afterwards, it will train a finetuned trajectory predictor for 400 epochs with the learning rate of 1e-4. We directly follow the default training strategy stated in the official code package for the network training.

\textbf{Dataset Preprocssing: } We use the provided preprocess code to generate rasterization of map images for each data sample. For the behavior probability maps, we use the equation:

\begin{equation}
    \mathcal{P}^{(i,p)}_{\mathcal{B}}(x,y) = \frac{\mathcal{P}^{(i,p)}_{\mathcal{B}}(x,y)}{\max(\mathcal{P}^{(i,p)}_{\mathcal{B}}(x,y))},
\end{equation}
which is stated in the paper to generate each probability map.

\textbf{Network Architecture: } P2T uses reinforcement learning to solve the prediction problem. Since we only modify the scene encoder part, in Figure \ref{fig:p2t}, we only shows the detailed modification in the encoder part and skip the description of the reward model and the decoder. 
\begin{figure}[tbh!]
    \centering
    \includegraphics[width=0.8\textwidth]{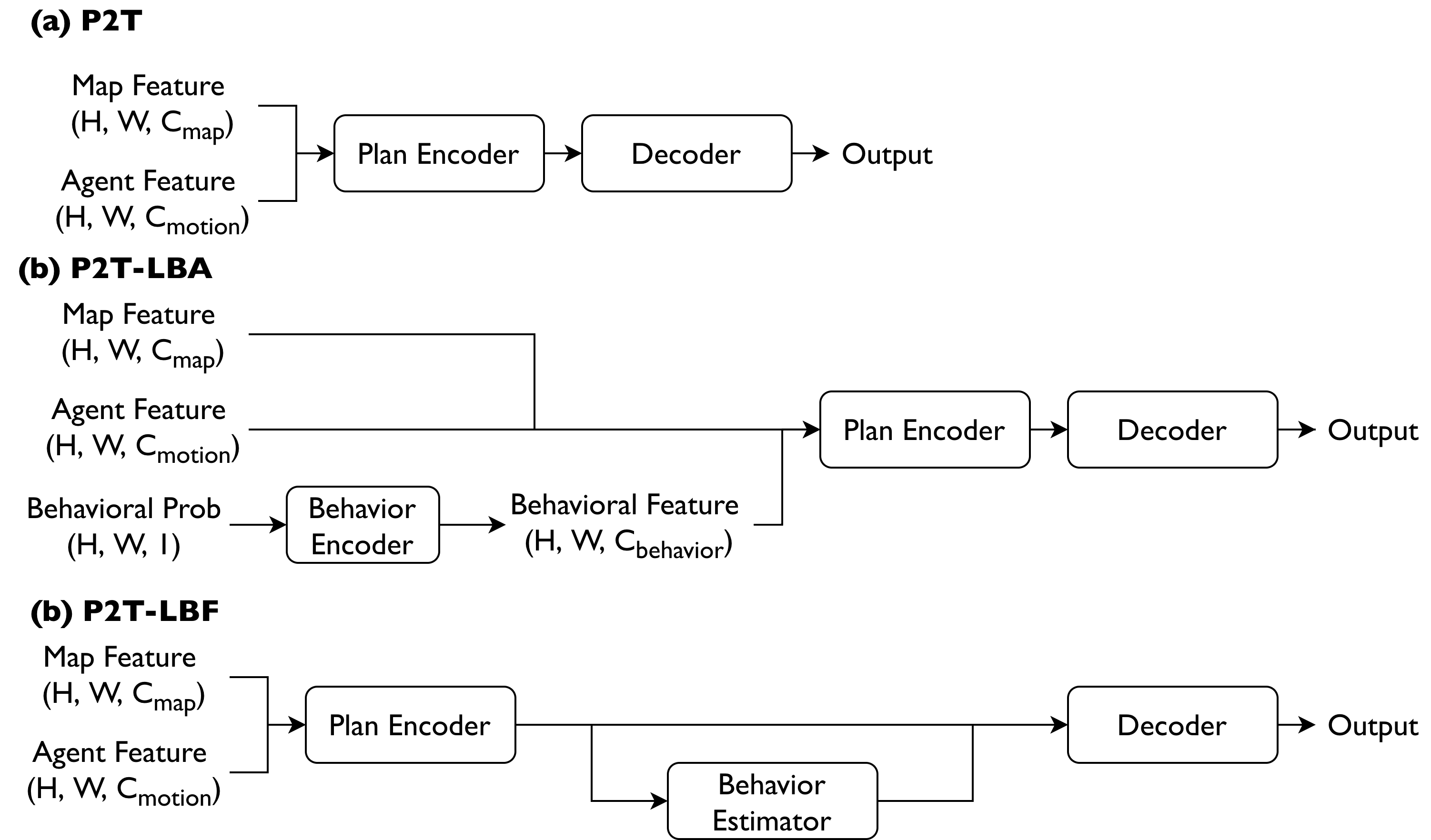}
    \caption{Implementation of P2T}
    \label{fig:p2t}
\end{figure}
The behavior encoder is one linear layer with activation and the output channel is 16. The behavior estimator is implemented as stacked 3-layer-conv2D structure, the kernel size is 1 and the output feature channel is 16. During the training of LBF framework, we apply KD loss on the output of the behavior estimator as well as the input of the Decoder.

\section{Additional Experiments}

\subsection{Ablation study on LBF system design.}
The LBF system has three factors: \textit{if using} \textbf{i}) knowledge distillation (KD), \textbf{ii}) LBA as the teacher net in KD (to provide local behavior features guidance), \textbf{iii}) behavior estimator. Tab \ref{table:abl} studies each factor for LaneGCN on Argoverse, and it shows: i) A surpassed baseline (self-KD is effective); ii) B surpassed A (local behavior features from LBA network are effective), iii) C surpassed B (the behavior estimator is effective).

\begin{table}[tbh!]

\setlength\tabcolsep{1.5pt}
    \centering
     \begin{tabular}{c||c|c||l|l}
    
\hline
Method  &   KD   &Behavior Estimator    &    minADE$_{1}$ &  minFDE$_{1}$\\
\hline
baseline &  no teacher    &   &   1.71 -   &   3.78 - \\
A&   self-KD  &   &   1.66\textcolor{red}{$\downarrow3\%$}    &   3.67\textcolor{red}{$\downarrow3\%$}\\
B&   LBA as teacher  &   & 1.65\textcolor{red}{$\downarrow3\%$}    & 3.63\textcolor{red}{$\downarrow4\%$}\\
C (LBF) &  LBA as teacher   &\checkmark   & 1.60\textcolor{red}{$\downarrow6\%$}    & 3.54\textcolor{red}{$\downarrow6\%$}\\

\hline
\end{tabular}
\caption{ {\small Each of three factors counts validated on Argo test split}}
\label{table:abl}
\end{table}

\subsection{Comparison with memory-based method.}
Table \ref{table:mantra} shows that the proposed LBA/ LBF significantly outperform SOTA memory-based method, MANTRA\cite{marchetti2020mantra}, on Argoverse test split (numbers from its original paper). As mentioned in Sec 2, previous memory-based methods also leverage historical info, but three major differences mark the novelty of our work: \textbf{i}) we directly use historical trajectories as system input to avoid information loss, \textbf{ii}) we explicitly emphasize the spatial locality, {\textbf{iii}) based on knowledge distillation, LBF does not need extra input while a memory-based method needs to store a huge memory bank.} 
\begin{table}[tbh!]
\setlength\tabcolsep{1.5pt}

    \centering
    \begin{tabular}{c|l|l|l|l}
\hline
Method  &   minADE$_{1}\downarrow$ & minFDE$_{1}\downarrow$ & minADE$_{6}\downarrow$ & minFDE$_{6}\downarrow$	\\

\hline
\rule{0pt}{1.6ex}
MANTRA&	2.36 -	&	5.31 - 	&	1.22 - 	&	2.30 -   \\
LaneGCN-LBA (ours)   &   \textbf{1.62 \textcolor{red}{$\downarrow30\%$}}  &   \textbf{3.58 \textcolor{red}{$\downarrow33\%$}} &  \textbf{0.84 \textcolor{red}{$\downarrow31\%$}} &   \textbf{1.30 \textcolor{red}{$\downarrow43\%$}} \\
LaneGCN-LBF (ours) 	&	\textbf{1.60 \textcolor{red}{$\downarrow32\%$}}	&	\textbf{3.54 \textcolor{red}{$\downarrow33\%$}}	&	\textbf{0.85 \textcolor{red}{$\downarrow30\%$}}	&	\textbf{1.31 \textcolor{red}{$\downarrow43\%$}}\\
\hline
\end{tabular}
\caption{ { Proposed LBA and LBF systems outperform MANTRA.}}
\label{table:mantra}
\end{table}

\subsection{Performance gains along the time}
In our paper, we only use the a few second local behavior data to avoid data snooping and for fair comparison. But in practice, by ``local," local behavior data just means the data's starting position is at an agent's current location; such data can be a long trajectory reflecting long-term behavior. We want to demonstrate the potential of the local behavior data for long-term prediction in real world application by using fig \ref{fig:time} to show that more gain from local behavior data as prediction time goes on. One reason is that as real human behavior recordings, local behavior data can suppress error accumulation in a prediction model.
\begin{figure}[tbh!]
\setcounter{figure}{7}

\centering
\includegraphics[width=0.8\textwidth]{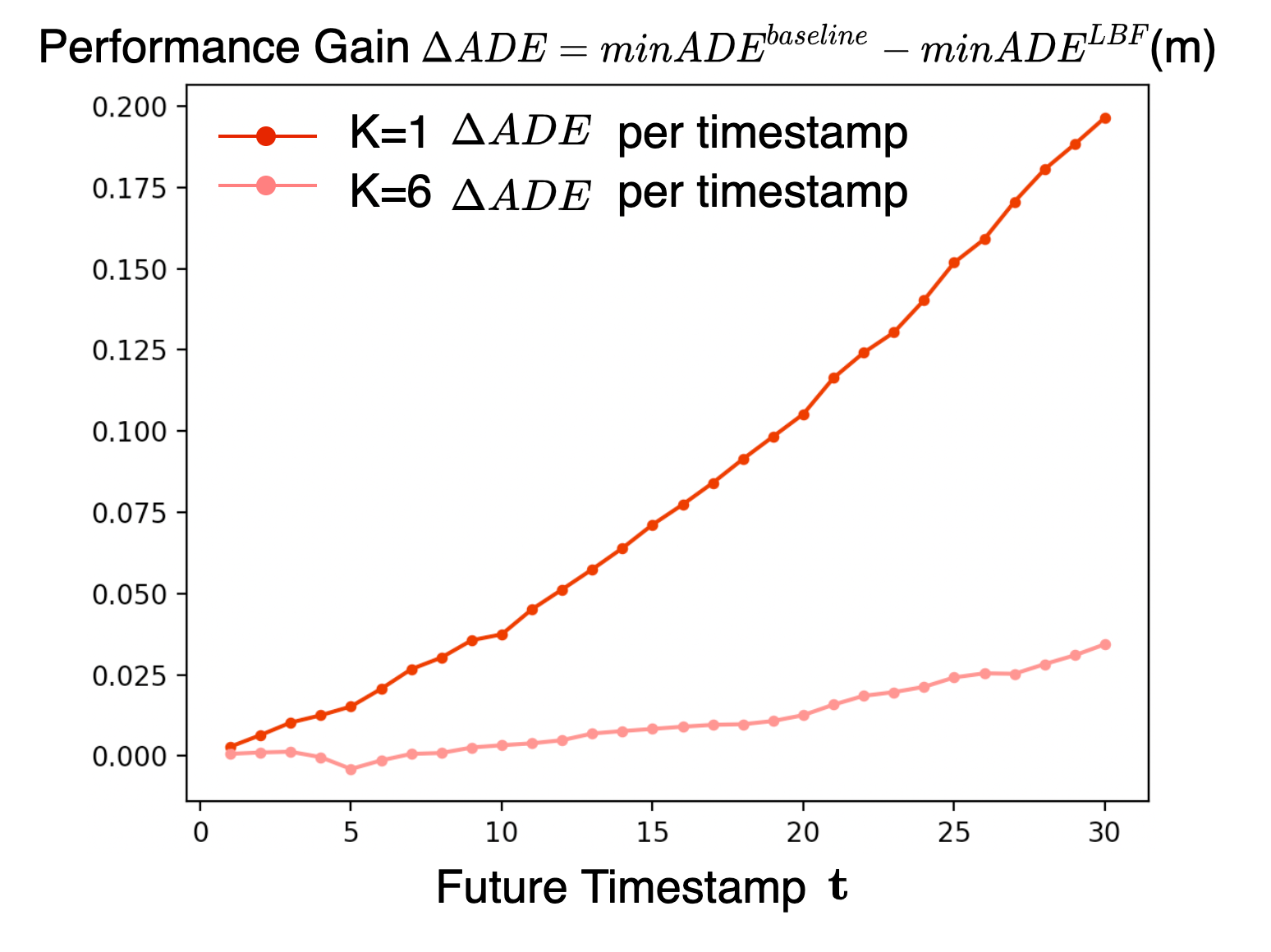}
\caption{Performance gain from LaneGCN-LBF compared to the baseline LaneGCN on Argoverse val set}
\label{fig:time}
\end{figure}

\end{document}